\documentclass[sigconf]{acmart}
\usepackage{times}
\usepackage{graphicx}
\usepackage{amsfonts}
\usepackage{amssymb}
\usepackage{amsmath}
\usepackage{bm}
\usepackage{graphicx}
\usepackage{subfigure}
\usepackage{bbm}

\def\BibTeX{{\rm B\kern-.05em{\sc i\kern-.025em b}\kern-.08emT\kern-.1667em\lower.7ex\hbox{E}\kern-.125emX}}

\title{Adversarial Multi-Binary Neural Network for Multi-class Classification 
}

\begin{document}

\title{Adversarial Multi-Binary Neural Network for Multi-class Classification 
}
\author{Haiyang Xu}
\affiliation{%
  \institution{AI Labs, Didi Chuxing}
  \city{Beijing}
  \country{China}}
\email{xuhaiyangsnow@didiglobal.com}

\author{Junwen Chen}
\affiliation{%
  \institution{AI Labs, Didi Chuxing}
  \city{Beijing}
  \country{China}}
\email{chenjunwen@didiglobal.com}

\author{Kun Han}
\affiliation{%
  \institution{AI Labs, Didi Research America}
  \city{Mountain View}
  \country{United States}}
\email{kunhan@didiglobal.com}

\author{Xiangang Li}
\affiliation{%
  \institution{AI Labs, Didi Chuxing}
  \city{Beijing}
  \country{China}}
\email{lixiangang@didiglobal.com}

\begin{abstract}
Multi-class text classification is one of the key problems in machine learning and natural language processing. Emerging neural networks deal with the problem using a multi-output softmax layer and achieve substantial progress, but they do not explicitly learn the correlation among classes. In this paper, we use a multi-task framework to address multi-class classification, where a multi-class classifier and multiple binary classifiers are trained together. Moreover, we employ adversarial training to distinguish the class-specific features and the class-agnostic features. The model benefits from better feature representation. We conduct experiments on two large-scale multi-class text classification tasks and demonstrate that the proposed architecture outperforms baseline approaches.
\end{abstract}
\maketitle
\section{Introduction}
Multi-class classification is a classic task for machine learning. One need to assign a label for a given example, where the number of possible labels are more than two. This is a very common problem in many areas, including natural language processing (NLP), computer vision, etc. 


Recent text classification utilizes neural networks. If the last layer uses a softmax function, it would be straightforward to deal with multi-class problem. However, this structure treats all class independently and neglects the relations among classes. From the perspective of representation learning, it does not guarantee that the raw input feature contributes equally to each class. Furthermore, some information in the raw feature may not be useful for the classification and compromise the performance.

Previous studies proposed using the one-vs-rest (OVR) scheme to deal with the multi-class problem \cite{rifkin2004defense,hastie2009multi,galar2011overview}. In neural networks, one can build multiple OVR classifiers in one network, where the top layers conduct independent OVR tasks and the bottom layers are shared among different classifiers. This structure is inspired by multi-task learning \cite{pan2010survey}, which aims to learn the correlation between related tasks to improve classification by learning them jointly. Multi-task learning learns the feature from different aspects and potentially regularizes the model to achieve better generalization.

In this paper, we treat each OVR binary classification as a single task, and all OVR binary classification together with the original multi-class classification constitute a multi-task problem. The rationale is that, for multi-class classification, the raw feature from the input space usually contains both information shared by all classes (class-agnostic) and specific to each class (class-specific). The former is shared among all classes and provide little information for classification, while the latter contains critical information for each class. To image that, in text classification, most stop words do not contribute to classification, which can be considered as class-agnostic, while the model should pay attention to other meaningful words.

To enforce the model to separate the features, we incorporate an adversarial training strategy. Specifically, each OVR binary classifier has a class-specific feature extractor, and we use another feature extractor to generate the class-agnostic feature which is then fed into a discriminator. The goal of the discriminator is to determine the source of the incoming feature, while the goal of the generator is to extract class-agnostic feature to fool the discriminator. Note that, in this paper, we use this architecture to address text classification problems, but it is straightforward to extend the proposed architecture to other domains.

\section{Related Work}
Multi-class classification has been studying for decades. Researchers designed one-vs-all and one-vs-one schemes and utilized many machine learning approaches to address the problem, including support vector machines  \cite{rifkin2004defense}, Adaboost \cite{hastie2009multi}, decision trees \cite{galar2011overview}, etc. Recently, neural networks are commonly used for classification. When utilizing the softmax layer, there is no essential difference between the binary and multi-class classification. 

In neural network based text classification, convolutional neural networks (CNNs) have been widely used to extract word-level or character-level feature representations \cite{collobert2011natural,kim2014convolutional,zhang2015character,conneau2017very}. Recurrent neural networks directly encode sequential structures and are suitable for text classification \cite{tai2015improved,lai2015recurrent,tang2015document,yang2016hierarchical}. Most recent trends suggest strong learning capacity of transformers with the attention mechanism and have shown impressive results on many NLP tasks \cite{vaswani2017attention,radford2018improving,devlin2018bert}. We mention that our approach focuses on a new multi-class training mechanism which is compatible with any type of text neural encoders. We choose the hierarchical attention network (HAN) \cite{yang2016hierarchical} to encode the text in documents in our experiments.

Our work also relates to multi-task learning and adversarial learning. \citet{collobert2008unified} proposed a unified multi-task framework for NLP. Many multi-task learning designed encoding network to learn shared information among different tasks \cite{luong2015multi,liu2016recurrent,sogaard2016deep,peng2017deep,subramanian2018learning}.  \citet{clark2018semi} proposed a cross-view method to exact features from different views of the neural networks. Adversarial learning has been gaining increasing attention since \citet{goodfellow2014generative}. \citet{miyato2016adversarial} proposed to use adversarial perturbations for text classification. \citet{chen2018adversarial} used adversarial networks to transfer the knowledge learned from labeled data for low resource classification. \citet{liu2017adversarial} adopted adversarial training to distinguish shared and private features for multi-task learning.

\section{Methodology}
\begin{figure}[h]
	\centering
	\includegraphics[width=0.9\linewidth]{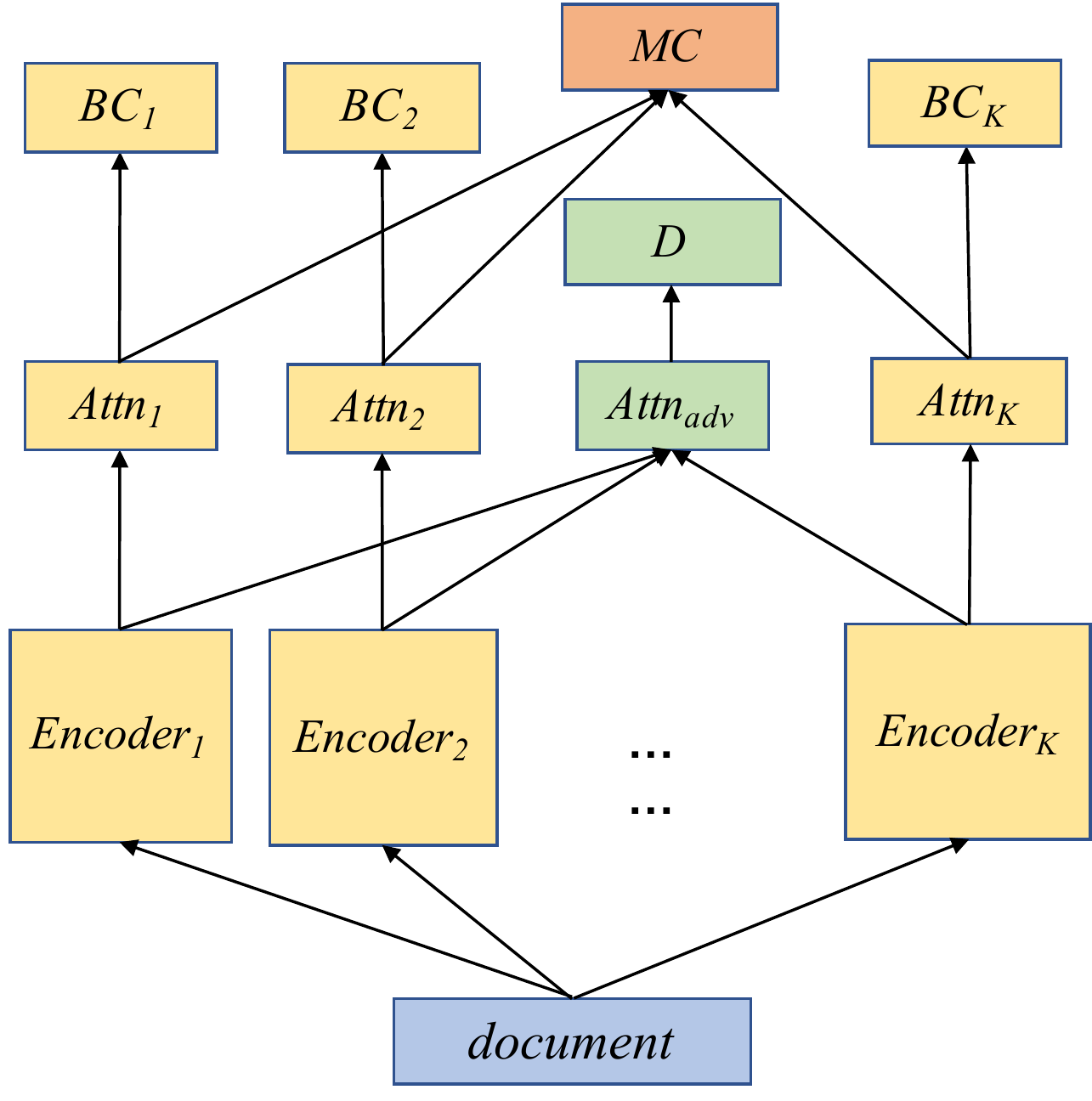}
	\caption{The architecture of our proposed model. $BC_i, Attn_i, Encoder_i$ are corresponding to the $i$th OVR binary classifier, and $MC$ denotes the multi-class classifier. $D$ is the discriminator and $Attn_{adv}$ is the adversarial attention}
	\label{fig:adv_binary}
\end{figure}

\subsection{Problem Formulation}
Each document $\bm d_i \in D$ consists of a sequence of sentences $<\bm s_1, \bm s_2, ...>$ and each sentence is corresponding to a sequence of tokens. Given a set of documents $D$, the  $K$-class text classification can be formulated as a mapping $f: D \rightarrow{} \{l_1,...,l_K\}$. We decompose the multi-class classification into $K$ binary classifications by the OVR strategy, so each binary classifier is $f_k: D \rightarrow{} \{l_k, l_{-k}\}$, where $l_k$ denotes the $k$th class and $l_{-k}$ denotes the superclass containing all classes except the $k$th class.

\subsection{Adversarial Multi-binary Neural Network}
As shown in Figure \ref{fig:adv_binary}, for a $K$-class classification problem, we build an encoding neural network for each OVR binary classification $f_k$. We use the hierarchical attention network (HAN) \cite{yang2016hierarchical} as an encoder for document classification in this study. The output of each HAN encoder is an attention module, which is then fully connected to a binary classifier with a two-output softmax layer. Formally, given a document $\bm d$, for the $k$th class, we use an encoder with an attention module to compute its class-specific representation $\bm a_k$ for binary classification:
\begin{align}
    &\bm e_k=\text{Encoder}_k(\bm d; \theta_{e_k}) \\
    &\bm a_k=\text{Attn}_k(\bm e_k; \theta_{a_k}) \\
    &P_{\text{bin}_k}(k|\bm d)=\frac{\exp({\bm w_k^T \bm a_k})}{\exp({\bm w_k^T \bm a_k}) + \exp({\bm w'{}_k^T \bm a_k})}
\end{align}

Here, $\text{Encoder}_k$ and $\text{Attn}_k$ are neural networks for the $k$th class with trainable parameters $\theta_{e_k}$ and $\theta_{a_k}$, which convert a document input into a fixed-length feature vector $\bm a_k$. $\bm w_k$ and $\bm w'_k$ are two weight vectors in the fully-connect layer corresponding to two softmax outputs. We use the negative log likelihood as the loss function for the $k$th binary classifier:
\begin{align}
    L_{\text{bin}_k}=-(y_k\log P_{\text{bin}_k}(k) + (1-y_k)\log P_{\text{bin}_k}(-k))
\end{align}
where $y_k$ is the ground-truth label for the $k$th classifier which is $1$ if $y_k=k$ otherwise 0. $P_{\text{bin}_k}(-k)$ is the probability of the softmax output not corresponding to $k$.

Besides, we take all $K$ class-specific feature vectors $\bm a_k$ as inputs to perform standard multi-class classification with multi-output softmax.
\begin{align}
    P_{\text{mul}}(k| \bm d)=\frac{\exp({\bm u_k^T \bm a_k})}{\sum_{i=1}^K\exp({\bm u_i^T \bm a_i})}
\end{align}
where $\bm u_i$ denotes the weight vector between the $i$th attention module to the multi-class softmax layer. Therefore, the loss function is:
\begin{align}
    L_{\text{mul}}=-\sum_{k=1}^K y \log P_{\text{mul}}(k)
\end{align}

To distinguish the class-specific and class-agnostic features, we employ an adversarial training structure including an adversarial attention model $\text{Attn}_{adv}$ and a discriminator $D$. For each training example, $\text{Attn}_{adv}$ takes as an input from each class-specific encoder and generate $K$ adversarial training instances. The goal of the discriminator is to determine which \textit{class-specific encoder} the current instance comes from, regardless which \textit{ground-truth class} the current example belongs to. The adversarial attention $\text{Attn}_{adv}$ serves as a generator aiming to extract class-agnostic features to fool the discriminator $D$:
\begin{align}
    &\bm a_{adv}(k)=\text{Attn}_{adv}(\bm e_k; \theta_{adv})  \\
	&P_D(j|k)= \frac{\exp((\bm v_{j}\bm a_{adv}(k))}{\sum_{i=1}^K \exp(\bm v_i \bm a_{adv}(k)) }	\\
	&L_{\text{adv}} = \min_{\theta_{adv}}(\lambda \max_{\theta_{D}}(\sum_{k=1}^{K}\sum_{j=1}^{K}z_k^j\log(P_D(j|k))) 
\end{align}
where $\bm a_{adv}(k)$ is the attention computed from the $k$th encoder. $z_k^j$ is the training target for $D$ which is equal to $1$ if $k=j$ otherwise 0. $\bm v_{i}$ is the weight vector for the discriminator $D$ and $\theta_D=\{\bm v_i; i=1,...,K\}$. $\lambda$ is a hyper-parameters.

The model is trained to extract class-specific features and class-agnostic features separately. The feature extracted from the each binary classifier should contribute more to the corresponding class than others, and we will demonstrate this in the experiment section. Furthermore, we apply orthogonality constraints \cite{liu2017adversarial} to prevent class-specific features from containing class-agnostic features. In the inference stage, we only take the class-specific features from the multi-class classifier as the final prediction. 

Finally, the training loss of the whole model is the summation of all loss:

\begin{align}
    L_{\text{diff}} = \sum_{k=1}^K\|{\bm a_k}^T\bm a_{adv}(k)\|_F^2
\end{align}

\begin{align}
    L=\alpha \sum_{k=1}^K L_{\text{bin}_k} + \beta L_{\text{mul}} + \gamma L_{\text{adv}} + \delta L_{\text{diff}}
\end{align}

where $\alpha$, $\beta$, $\gamma$ and $\delta$ are hyper-parameters.
\section{Experiments}
\subsection{Datasets and Baselines}
To evaluate the proposed approach, we apply two large scale multi-class document datasets. 



\begin{itemize}
\item \textbf{IMDB reviews} \cite{diao2014jointly}: it is formed by randomly select 50k movies and crawl all their reviews from IMDb. There are around 348,000 documents in the dataset and 10 classes in total. We randomly select 80\% of the data for training, 10\% for validation and the remaining 10\% for test.

\item \textbf{Yahoo answers} \cite{zhang2015character}: it is a topic classification task with 10 topic classes, which includes 140,000 training samples and 5,000 testing samples. We randomly select 10\% of the training samples as validation set.

\end{itemize}

For comparison, we choose the following methods based on deep learning as the baseline approaches:
\begin{itemize}
\item \textbf{CNN}:\citet{kim2014convolutional} take the whole document as a single sequence and uses a CNN with pooling for classification (CNN). \item \textbf{Conv-GRNN and LSTM-GRNN }:\citet{tang2015document} use a CNN or LSTM to form a sentence vector and then use a gated recurrent neural to combine the sentence vectors to a document level vector representation for classification 
\item \textbf{HAN}: \citet{yang2016hierarchical} use a hierarchical attention network to model the sentences and the document with different pooling methods (HAN-\{ATT, AVE, MAX\}).
\end{itemize}




\subsection{Implementation}
We take the same pretrain mechanism as in \cite{yang2016hierarchical}, which retain words appearing more than 5 times in the vocabulary and obtain the word embedding using word2vec \cite{mikolov2013distributed}. Furthermore, the parameters of our model are the following: we set the word embedding dimension to be 200 and the BiLSTM hidden state dimension to be 100. The word/sentence context vectors dimension also are set to be 100. We use RMSProp with learning rate 0.001. The hyper-parameters are tuned on the validation set. The $\alpha$ is set as 0.5, $\beta$ is set as 1, $\gamma$ is set as 0.1 and $\delta$ is set as 0.1.  

\subsection{Results and analysis}
The experimental results on two datasets are shown in Table \ref{tab:res}. We compare the baseline approaches with our adversarial multi-binary approach (AMB) in terms of classification accuracy. In order to evaluate the effectiveness of the adversarial training, we also train multi-binary models without adversarial training (MB). Results of baselines are taken from the corresponding papers. From the results, we can observe that: (1) Our AMB and MB models outperform the baselines on both datasets (+1.4\% and +2.5\% on IMDB, +1.0\% and +1.7\% on Yahoo), showing the advantage of the multi-binary scheme in multi-class problem. The results are consistent in different classification tasks. (2) Comparing MB and AMB, the adversarial training further boosts the performance, indicating that better feature representation is learned from the adversarial model.

\begin{table} 
\caption{The experimental results on IMDB reviews and Yahoo answers datasets, in percentage. MB stands for multi-binary model and AMB stands for adversarial multi-binary model.}
\centering
       \begin{tabular}{c c c c}
       \hline
    Methods & IMDB & Yahoo  \\\hline \hline
    CNN & 34.1 & 71.2  \\
    Conv-GRNN & 42.5 & - \\
    LSTM-GRNN & 45.3 & - \\
    HAN-AVE & 47.8 & 75.2 \\
    HAN-MAX & 48.2 & 75.2 \\
    HAN-ATT & 49.4 & 75.8 \\ \hline \hline
    \textbf {MB} & 50.8 & 76.8 \\
    \textbf {AMB} & \textbf {51.9} & \textbf{77.5} \\ \hline
    \\
    \end{tabular}
       \label{tab:res}
\end{table}

\begin{figure}[h]
	\centering
	\includegraphics[width=2.5in,height=3.5in]{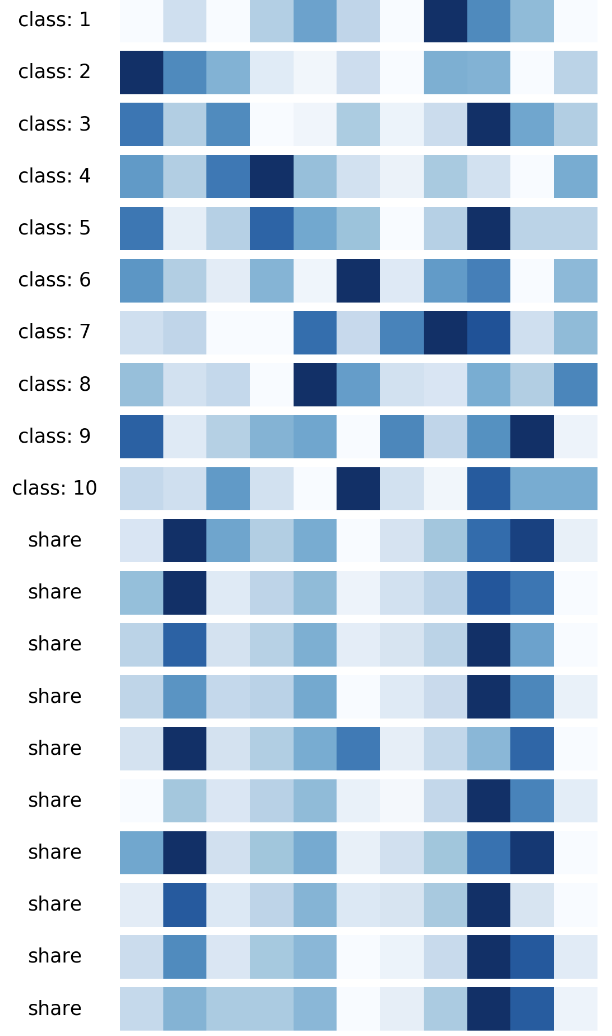}
	\caption{The example of attention weight distribution of each binary subtask in dataset Yahoo}
	\label{fig:attention}
\end{figure}

\subsection{Visualization}
Our model aims to separate class-specific features and class-agnostic features. We randomly choose a document from Yahoo dataset and visualize the corresponding attention to validate the idea. In Figure \ref{fig:attention}, each row represents the attention strength over all sentences in this document, and the depth of color represents the strength of the weight. As shown in the figure, the upper 10 rows are the class-specific attention ($\bm a_k$) which have different activation pattern. The lower 10 rows are corresponding to the adversarial attention from each class-specific encoder ($\bm a_{adv}(k)$), and the patterns in the adversarial attention (shared) are very similar, suggesting that it learns the class-agnostic information.

\section{Conclusions}
 In this paper, we consider a multi-class problem as multiple binary classifications using a multi-task framework. We employ an adversarial training approach to learn the information shared and specific to each class, which learns better feature representation and improve the classification performance. Experiments show the superiority of the proposed approach.

  \bibliographystyle{ACM-Reference-Format}
  \bibliography{binary_text_classification}
\end{document}